\documentclass{article}

    \PassOptionsToPackage{numbers, compress}{natbib}

\usepackage[preprint]{neurips_fitml_2024}

\usepackage[utf8]{inputenc} % allow utf-8 input
\usepackage[T1]{fontenc}    % use 8-bit T1 fonts
\usepackage{hyperref}       % hyperlinks
\usepackage{url}            % simple URL typesetting
\usepackage{booktabs}       % professional-quality tables
\usepackage{amsfonts}       % blackboard math symbols
\usepackage{nicefrac}       % compact symbols for 1/2, etc.
\usepackage{microtype}      % microtypography
\usepackage{xcolor}         % colors
\usepackage{graphicx}
\usepackage{multirow}
\usepackage{wrapfig}
\usepackage{amsmath}
\usepackage{algorithm}
\usepackage{algpseudocode}

\title{Simultaneous Weight and Architecture Optimization for Neural Networks}

\author{%
  Zitong Huang \\
  Department of Electrical \& Computer Engineering\\
  University of Southern California\\
\AND
 Mansooreh Montazerin \\
  Department of Electrical \& Computer Engineering\\
  University of Southern California\\
\AND
 Ajitesh Srivastava \\
  Department of Electrical \& Computer Engineering\\
  University of Southern California\\
}

\begin{document}

\maketitle

\begin{abstract}
Neural networks are trained by choosing an architecture and training the parameters. The choice of architecture is often by trial and error or with Neural Architecture Search (NAS) methods. While NAS provides some automation, it often relies on discrete steps that optimize the architecture and then train the parameters. We introduce a novel neural network training framework that fundamentally transforms the process by learning architecture and parameters simultaneously with gradient descent. With the appropriate setting of the loss function, it can discover sparse and compact neural networks for given datasets. Central to our approach is a multi-scale encoder-decoder, in which the encoder embeds pairs of neural networks with similar functionalities close to each other (irrespective of their architectures and weights). To train a neural network with a given dataset, we randomly sample a neural network embedding in the embedding space and then perform gradient descent using our custom loss function, which incorporates a sparsity penalty to encourage compactness. The decoder generates a neural network corresponding to the embedding. Experiments demonstrate that our framework can discover sparse and compact neural networks maintaining a high performance.
\end{abstract}

\section{Introduction}
Nowadays, the growing use of Deep Learning (DL) to address complex Artificial Intelligence (AI) problems has made the selection of an appropriate neural network increasingly critical. In recent years, numerous neural architectures have emerged and been widely used in the AI field. There is a vast number of neural networks such as ResNet \cite{he2016deep} and VGG \cite{simonyan2014very} which achieved remarkable success in image processing tasks; similarly, architectures like Transformer \cite{vaswani2017attention} and BERT \cite{devlin2018bert} have revolutionized natural language processing. However, selecting and training an optimal neural network remains a time-consuming and experience-dependent process. To address this issue, scientists have introduced a new research area known as Neural Architecture Search (NAS), which aims to automate the process of designing neural networks \cite{wistuba2019survey}. It leverages computational methods to discover optimal network structures, enabling efficient exploration of a vast design space. Currently, the main NAS methods can be categorized into discrete and continuous search: reinforcement learning and evolutionary algorithms search \cite{zoph2016neural, liu2021survey} for the best architecture in a discrete search space, while continuous methods introduce gradient-based techniques to optimize neural architectures \cite{elsken2019neural, ren2021comprehensive}. 

Traditionally, methods such as reinforcement learning \cite{zoph2016neural, baker2016designing} and evolutionary algorithms \cite{liu2021survey} represent neural architectures using discrete high-dimensional representations. These methods involve more complex computations as they must search through a vast and discrete space of potential architectures. Furthermore, they are constrained by the predefined set of available operations, which limits the diversity of architectures that can be explored. As a result, continuous search approaches in NAS have gradually gained advantage, offering more flexibility by allowing smooth optimization over a continuous space. Scientists are exploring different methods to project neural architectures into continuous spaces~\cite{luo2018neural, li2020neural}. These methods encode architectures into a continuous embedding space and then perform optimal architecture search on that space through gradient descent. While gradient-based methods have gained significant attention due to their ability to perform differentiable optimization and to provide high computational efficiency, they search for the optimal architecture first and then, train it to obtain the optimal weights. This separation of architecture search and weight optimization steps is a cumbersome process and may overlook some promising architectures.

We propose a novel gradient-based neural network search method, aiming to \textit{simultaneously search for the optimal neural network architecture and weights within a continuous space}. We utilize a multi-scale encoder-decoder framework, in which the encoders push the neural networks with similar functionality closer together in the embedding space, and the decoders attempt to reconstruct the neural network from the embedding space. Finally, by performing gradient descent within the trained embedding space, we simultaneously optimize both the architecture and the weights. Having designed an efficient loss function for performing gradient descent, we achieve a high-performance neural network that is both sparse and compact. The experiments are conducted on Multilayer Perceptrons (MLPs) with 3 different activation functions, i.e., sigmoid, leaky-ReLU, and linear. Note that our objective is not NAS alone, but obtaining a model directly (treating architecture and its parameters as a single entity). To the best of our knowledge, our framework is the first to approach the training of neural networks in this way.
%Our method employs multiple encoders and decoders that determine the space over which we can perform gradient-based neural network search. 
Our contributions can be summarized as follows:
\begin{itemize}
    \item We propose a training method for simultaneously obtaining the optimal combination of architecture and weights using an autoencoder that embeds neural networks based on their functionality.
    \item We demonstrate that our method can train compact and sparse neural networks with various activations (sigmoid, leaky-ReLU, and linear).
\end{itemize}

% We will present a method overview in Section \ref{Method}, explaining how to train the autoencoder and utilize gradient descent to search for the optimal network from a given dataset. We then conduct experiments in Section \ref{Experiments} to validate the effectiveness of our method, demonstrating its capability to discover sparse and high-performing networks.

\section{Related Work}
\label{gen_inst}

One seminal work in discrete NAS is the NeuroEvolution of Augmenting Topologies (NEAT) algorithm \cite{stanley2002evolving} which combines evolutionary algorithms with neural network evolution and allows networks to grow in complexity over time. It has demonstrated its effectiveness in evolving neural network architectures for various tasks, including game-playing and robot control. At the same time, reinforcement learning-based approaches, such as the Proximal Policy Optimization (PPO), have also been applied to architecture search \cite{baker2016designing}. These methods learn to balance the trade-off between exploration and exploitation, efficiently navigating the architecture space to find networks that excel in specific tasks. A drawback of these approaches is that they rely on discrete search techniques, which can be inefficient and sub-optimal due to the large search space.

Differentiable Architecture Search (DARTS) has introduced gradient-based techniques to optimize neural architectures \cite{liu2018darts}, offering faster and more accessible architecture discovery. It allows the selection of an operation from the set of candidates by applying a softmax function over all possible options. The operation with the highest value is chosen. However, DARTS assumes a fixed set of architectural candidates (e.g., different convolution sizes) between a fixed number of layers, and it inherently follows a discrete search framework, as each subsequent operation is based on the previous one. Additionally, DARTS cannot simultaneously optimize both weights and architecture.

%here you should talk about the 3rd method and then, talk about yours and its benefits.

Another continuous NAS approach employs Graph Variational Autoencoders (Graph VAEs) as part of their framework \cite{li2020neural, chatzianastasis2021graph}. This framework integrates an encoder, a performance predictor, a complexity predictor, and a decoder. The encoder and decoder are part of the Graph VAE, responsible for mapping neural architectures to and from continuous latent representations. The performance and complexity predictors are regression models that predict both the architecture's performance and computational cost, which will ensure that the architectures discovered are high-performing and computationally efficient. However, this method is limited to searching architectures tailored for a specific dataset. If the dataset changes, a new framework needs to be trained from scratch to adapt. Additionally, it separates the process of finding architectures from training the weights, which makes the process more cumbersome.

\section{Method}
\label{Method}

Our current work focuses on MLP networks. The major innovation of our framework is the use of a multi-scale encoder-decoder method to simultaneously search for the optimal network architecture and weights, achieving a compact and sparse MLP network. 

\subsection{Embedding MLPs using Autoencoder}
\label{others}

% Our framework primarily consists of a 
The proposed multi-scale autoencoder, which employs multiple encoders and decoders, takes an array representing the MLP structure as input. The encoders embed this array into a low-dimensional space, and then the decoders reconstruct it. The autoencoder is trained so that encoders learn an embedding space where two MLPs are close to each other if they produce similar outputs for the same inputs. Each decoder is trained to regenerate the array that represents the MLP network from the embedding.

\paragraph{Matrix Representation of MLP Networks.}
\label{matrix representation}
We use a matrix to represent an MLP that has the same number of neurons in all the hidden layers. Later we extend this representation to MLPs with varying numbers of neurons per layer (see ``Varying Number of Neurons'').
If the input size of an MLP network is $i$, the hidden layer size is $n$, and the output size is $o$, then an MLP network can be represented as: 

\begin{equation}
[W_{i, n}, W_{n, n}, W_{n, n}, \dots, W_{n, o}]
\end{equation}

Here, $W_{i, n}$ represents the weight matrix connecting the input layer (with size $i$) to the first hidden layer (with size $n$). The concatenation of all these weight matrices forms a comprehensive representation of the entire MLP network, with padding applied in the column dimension as needed to account for the varying layer sizes.
% Their concatenation can be considered as a matrix representation of the MLP network, with some padding required in the column dimension.

\paragraph{Architecture of Multi-Scale Encoder-Decoder}
\label{multi-scale encoder-decoder}

% \begin{figure}[htp]
%     \centering
%     \includegraphics[width=8cm]{large_framework.png}
%     \caption{The overall framework for the multi-scale encoder-decoder}
%     \label{Whole_Framework}
% \end{figure}
\begin{wrapfigure}{r}{0.65\textwidth}
    \centering
    \includegraphics[width=\linewidth]{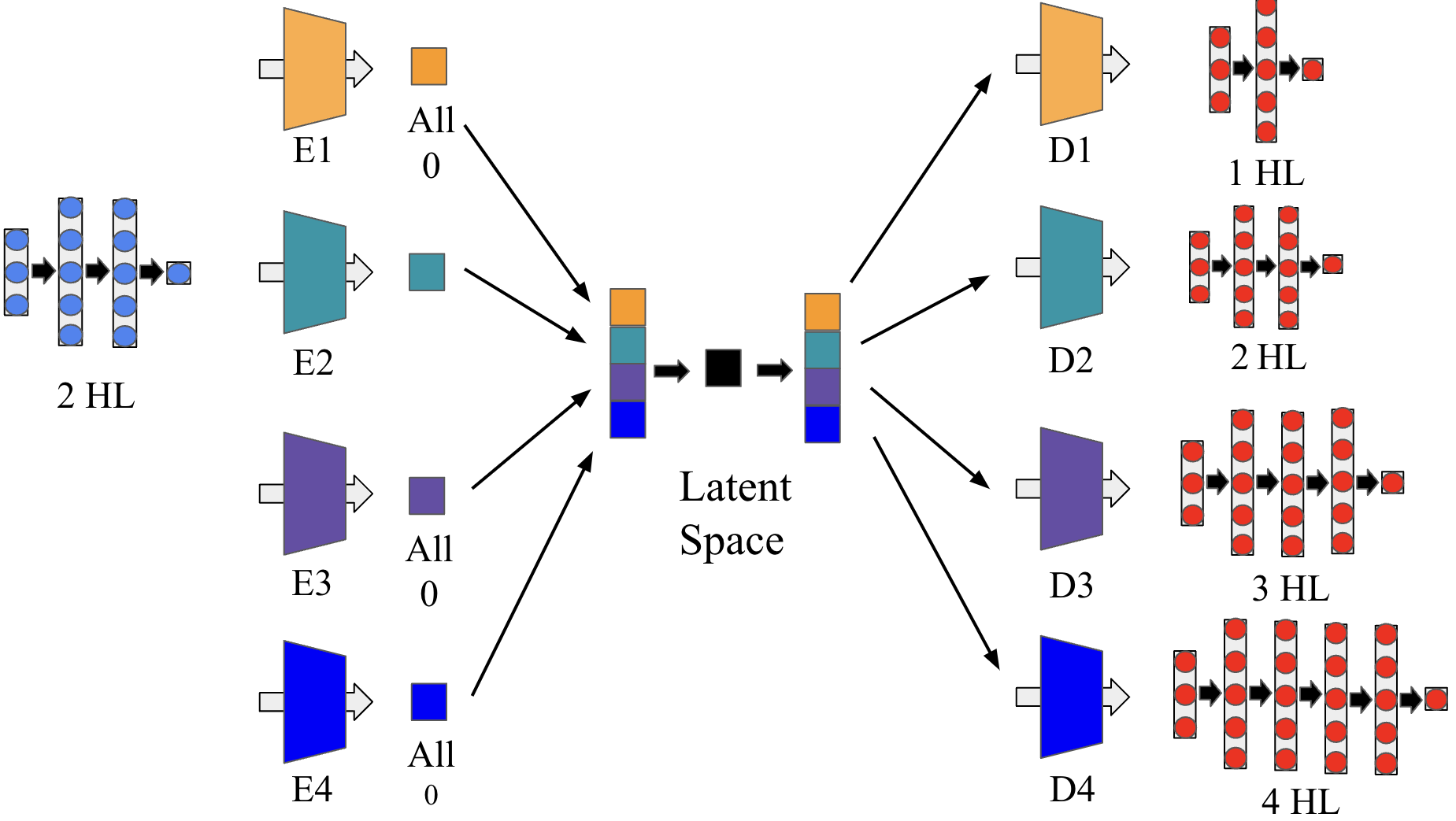}
    \caption{The overall framework for the multi-scale encoder-decoder.}
    \label{Whole_Framework}
\end{wrapfigure}
We use a multi-scale autoencoder to search for optimal MLP architectures with different numbers of hidden layers and their corresponding weights. %A similar approach has been used in the literature to generate images of multiple resolutions \cite{khan2021pmed}. 
Each encoder and decoder corresponds to an MLP with a specific number of hidden layers. The overall framework is illustrated in Figure~\ref{Whole_Framework}. Here, we encode and decode MLPs with 1, 2, 3, and 4 hidden layers.

The set of encoders ($E_1, E_2, \dots$) functions like a multiplexer: when we have an MLP with $i$ hidden layers in input, only encoder $E_i$ is activated to encode the input MLP, while all the other encoders output a vector of all zeros. Each encoder is composed of 7 convolutional layers. We concatenate the encoding results from all encoders, and then use 4 Fully Connected (FC) layers to embed them into a low-dimensional space. In this way, we project all MLP networks, regardless of their number of hidden layers, into the same embedding space.

During the decoding process, we first use 4 FC layers to transform the embedded vector back into a high-dimensional space. Then, depending on the number of decoders, denoted as $D$, we split this vector into multiple sub-vectors, each corresponding to the input of a specific decoder. This step acts like a switch, using some linear transformations to determine what input should be assigned to each decoder. Subsequently, a decoder $D_i$ serves as an approximate inverse of its corresponding encoder $E_i$ to reconstruct the MLP networks with $i$ hidden layer using the embedded representation from the continuous embedding space. 

Through this method, we find MLP networks $M_1$, $M_2$, ..., $M_l$ with $1$, $2$, ..., $l$ hidden layers, respectively, that function similarly to the input MLP network for the same input data (not necessarily reconstructing the input MLP).

%%%%%Mansooreh
\paragraph{Varying Number of Neurons}
\label{Varying Number of Neurons}

To allow each hidden layer of the input and output MLP architectures have arbitrary sizes, we add additional columns to the MLP's matrix representation to include information about the hidden layer sizes. 

For the matrix representation of the input MLPs, we assume that each hidden layer can have up to $n$ active neurons, and there are $i$ hidden layers in total. We use the rule demonstrated in "Matrix Representation of MLP Networks" to represent an MLP network, and we add additional $i$ columns at the end of the matrix, with each column corresponding to a collection of neurons in a hidden layer. These additional columns consist of integer values of $1$ and $0$, indicating which neurons in each hidden layer are active. The corresponding weights of the inactive neurons in each hidden layer are masked. Finally, we feed this new representation into the encoders, encoding both the architecture and weights of an MLP network into the embedding space.

We use the method described in "Architecture of Multi-Scale Encoder-Decoder" to decode, but apply a sigmoid function to the last $i$ columns of the decoded matrix. If the sigmoid output is greater than 0.5, we treat the element as 1, indicating that the neuron is active. Otherwise, we treat the element as 0, indicating that the neuron is inactive. We mask the corresponding weight elements in the matrices to be 0 and ensure their output values are 0 when input values are fed into this decoded MLP.

\paragraph{Training Multi-Scale Autoencoder}
\label{loss function definition}

We first generate 1,000 sets of 3-dimensional inputs $\mathbf{x} \in [-1, 1]^3$, and then feed them into input MLPs, which are also randomly generated, to obtain the corresponding ground truth outputs. The specific process of generating the dataset is discussed in Section \ref{multi-scale training}. These input-output pairs and MLPs form the core of our dataset, which we use to train the multi-scale autoencoder network with the following choices of loss functions:
% $$Loss_i = (\sum_x (\mathcal{N}_s(x) - [H_i(G(\mathcal{N}_s)](x) )^2)^{p}$$ $$ Loss = \sum_{\mathcal{N}_s}(\sum_{i}Loss_i)^{1/p}$$
\begin{equation}
        L_{ED} = \sum_{\mathcal{N}_s} \left( \sum_{i=0}^{l} \left( \sum_x \left( \mathcal{N}_s(x) - [D_i(E(\mathcal{N}_s))](x) \right)^2 \right)^{p} \right)^{1/p}
        \label{p loss}
\end{equation}

\begin{equation}
    L_{ED} = \sum_{\mathcal{N}_s} \min_{i \in \{1, 2, ... l\}} \left(\sum_x \left( \mathcal{N}_s(x) - [D_i(E(\mathcal{N}_s))](x) \right)^2\right)
    \label{min loss}
\end{equation}
Here, $[D_i(E(\mathcal{N}_s)](x)$ represents the output of the decoded MLP with $i$ hidden layers on input value $x$, and $\mathcal{N}_s$ refers to input MLP networks. 

The first choice (p-norm loss) is designed to smoothly aggregate the loss across all the decoders. For instance, with $p=2$, we expect all the decoders to generate MLPs with low losses. On the other hand, setting $p=-2$ encourages the overall loss to focus more on the decoder producing the lowest loss. 
The second choice (min-loss) explicitly chooses the decoder with the lowest loss. However, it may introduce non-differentiability.
Our experiments demonstrated that the best performance was obtained from the min-loss function.

%We will maintain all three loss functions in our experiments for comparison.

\subsection{Training MLPs in the Embedding Space}

We use the trained multi-scale encoders-decoders to search for the optimal MLPs for a given dataset. Our objective is to achieve a compact and sparse MLP that can effectively represent the dataset.

\paragraph{Training Optimal MLPs for a Dataset} 

We randomly sample an MLP embedding from the embedding space, which serves as the starting point for finding the optimal MLP. Each embedding $z$ encodes both the architecture and weight information of the MLP. Thus, training the optimal MLP involves simultaneously searching for the best combination of architecture and weights. The entire search process is conducted via gradient descent, where the loss function measures the discrepancy between the predictions of the current MLP, defined by the current $z$, and the target outputs. Since we have multiple decoders $D$, we need to perform $l$ parallel gradient descent processes, where $l$ represents the number of decoders. Each decoder is designed to search for an MLP network with $l$ hidden layers. The loss function and the gradient descent process for each search are defined as follows:

% \begin{align}
%     L_i &= \sum_{x, y} \left( \left[ H_i(z) \right](x) - y \right)^2 \\
%     \nabla_z L_i &= 2 \sum_{x, y} \left( \left[ H_i(z) \right](x) - y \right) \nabla_z \left[ H_i(z) \right](x)
% \end{align}

\begin{align}
    L_i &= \sum_{x, y} \left( \left[ D_i(z) \right](x) - y \right)^2, \quad
    \nabla_z L_i = 2 \sum_{x, y} \left( \left[ D_i(z) \right](x) - y \right) \nabla_z \left[ D_i(z) \right](x)
\end{align}
% \[
%     L_i = \sum_{x, y} \left( \left[ D_i(z) \right](x) - y \right)^2, \quad
%     \nabla_z L_i = 2 \sum_{x, y} \left( \left[ D_i(z) \right](x) - y \right) \nabla_z \left[ D_i(z) \right](x)
% \]

In this context, $\left[ D_i(z) \right](x)$ represents the prediction values produced by the decoded MLP $D_i(z)$  for the input values
$x$, and $y$ refers to the ground truth output values. By utilizing these functions for gradient descent, we iteratively refine the continuous representation of the MLP to improve its performance on the dataset.

\paragraph{Sparsity and Compactness}

To enable the search for a more sparse and compact MLP, we introduce a sparsity penalty $S(z)$ in the loss function, which penalizes unnecessary complexity and encourages the elimination of redundant components. The specific loss function is defined as follows:

% \begin{equation}
%     L_i = \sum_{x, y} \left( \left[ H_i(z) \right](x) - y \right)^2 + S_i(z)
% \end{equation}

% \begin{equation}
%     S_i(z) = \alpha \left( \mathcal{L}_1 + \text{SoftCount} \right)
% \end{equation}

% \begin{equation}
%     \begin{aligned}
%         \mathcal{L}_1 &= 0.1 \|W[H_i(z)]\|_1 \\
%         \text{SoftCount} &= 0.5  \sigma \left( 10 \|W[H_i(z)] - t \|_1 \right)
%     \end{aligned}
% \end{equation}

\begin{equation}
L_i = \sum_{x, y} \left( \left[ D_i(z) \right](x) - y \right)^2 + S_i(z), \quad S_i(z) = \alpha \left( \mathcal{L}_1 + \text{SoftCount} \right)
\end{equation}

\begin{equation}
\mathcal{L}_1 = 0.1 \|W[D_i(z)]\|_1, \quad \text{SoftCount} = 0.5  \sigma \left( 10 \|W[D_i(z)] - t \|_1 \right)
\end{equation}

% \[
% L_i = \sum_{x, y} \left( \left[ D_i(z) \right](x) - y \right)^2 + S_i(z), \quad S_i(z) = \alpha \left( \mathcal{L}_1 + \text{SoftCount} \right)
% \]

% \[
% \mathcal{L}_1 = 0.1 \|W[D_i(z)]\|_1, \quad \text{SoftCount} = 0.5  \sigma \left( 10 \|W[D_i(z)] - t \|_1 \right)
% \]

The sparsity penalty consists of both L1 regularization and a soft counting switch. The purpose of L1 regularization is to minimize the absolute values of all weights. In the soft counting switch, we leverage the sigmoid function to model how many elements fall below a threshold $t$ and set them to zero. By introducing $t$ as a learnable parameter, we provide the loss function with more flexibility, as it creates additional pathways for adjustment. The hyperparameter $\alpha$ controls the weight of the sparsity penalty. The overall sparsity penalty encourages the MLPs to have smaller weights and fewer neurons, with more elements falling below the threshold $t$, thereby increasing sparsity.

\section{Experiments}
\label{Experiments}

% \label{multi-scale training}
% \begin{wraptable}{r}{0.4\textwidth}
% \vspace{-0.5in}
%   \caption{MPE results for multi-scale autoencoder using min-loss.}
%   \label{MPE_combined_table}
%   \centering
%     \begin{tabular}{lcccc}
%         \toprule
%         \multirow{2}{*}{Decoders} & \multicolumn{4}{c}{MIN Loss} \\
%         \cmidrule(lr){2-5} 
%         & $E_1$ & $E_2$ & $E_3$ & $E_4$ \\
%         \midrule
%         $D_1$ & \textbf{5.84} & 38.3 & 30.4 & 29.7  \\
%         $D_2$ & 16.6 & 35.9 & 36.9 & 45.7  \\
%         $D_3$ & 25.4 & \textbf{23.7} & \textbf{24.1} & \textbf{24.4} \\
%         $D_4$ & 8.37 & 27.5 & 36.0 & 44.7  \\
%         \bottomrule
%     \end{tabular}
% \end{wraptable}
% Our experiments are divided into two parts. We first test the effectiveness of the multi-scale encoder-decoder training and then evaluate the ability to search for optimal MLPs given a specific dataset. 

\label{multi-scale training}
\begin{table}
  \caption{MPE results for multi-scale autoencoder trained using three different loss functions}
  \label{MPE_combined_table}
  \centering
    \begin{tabular}{lcccc|cccc|cccc}
        \toprule
        \multirow{2}{*}{Decoders} & \multicolumn{4}{c|}{MIN Loss} & \multicolumn{4}{c|}{$P=-2$ Loss} & \multicolumn{4}{c}{$P=2$ Loss} \\
        \cmidrule(lr){2-5} \cmidrule(lr){6-9} \cmidrule(lr){10-13}
        & $E_1$ & $E_2$ & $E_3$ & $E_4$ & $E_1$ & $E_2$ & $E_3$ & $E_4$ & $E_1$ & $E_2$ & $E_3$ & $E_4$ \\
        \midrule
        $D_1$ & \textbf{5.84} & 38.3 & 30.4 & 29.7 & \textbf{8.93} & 34.4 & 36.0 & 43.4 & 31.4 & 38.4 & 37.8 & 45.1 \\
        $D_2$ & 16.6 & 35.9 & 36.9 & 45.7 & 16.7 & 30.5 & 42.0 & 45.6 & 40.8 & 37.7 & 34.1 & 36.2 \\
        $D_3$ & 25.4 & \textbf{23.7} & \textbf{24.1} & \textbf{24.4} & 13.1 & \textbf{28.1} & \textbf{25.6} & \textbf{27.3} & \textbf{27.2} & \textbf{35.5} & \textbf{33.1} & \textbf{35.2} \\
        $D_4$ & 8.37 & 27.5 & 36.0 & 44.7 & 53.4 & 28.5 & 35.1 & 43.2 & 35.0 & 39.9 & 36.2 & 38.2 \\
        \bottomrule
    \end{tabular}
\end{table}

\subsection{Multi-Scale Encoder-Decoder}

We encode sigmoid-based (with a linear activation used between the final hidden layer and the output layer), as well as leaky ReLU-based and linear-based (both with a sigmoid activation used between the final hidden layer and the output layer) MLPs with 1, 2, 3, and 4 hidden layers. The number of neurons in each hidden layer ranges from 3 to 7. We generate the MLPs and input values ourselves, and use them to produce the corresponding ground truth outputs, which serve as the dataset for training our multi-scale autoencoder. The process for generating the MLPs is discussed in detail in the Appendix. For each randomly generated MLP, we uniformly generate 3-dimensional input value combinations by sampling 10 values for each of the three dimensions, with each value ranging between -1 and 1. These combinations result in a total of 1,000 input sets. After feeding these inputs into the MLP network, we obtain 1,000 corresponding outputs. During training, we feed these MLPs into the multi-scale encoder-decoder and train them using the three loss functions described in Section \ref{loss function definition}. Our entire dataset consists of 3,200,000 MLPs, and for each training process based on the different loss functions, we train for 4 epochs. The entire training process for each autoencoder takes approximately 48 hours using an NVIDIA RTX A5000 GPU.

We then examine the effectiveness of the multi-scale autoencoder after training. We fix all parameters in the encoder and decoder framework. For each MLP with a specific number of hidden layers, we randomly generate 1,000 MLPs and calculate the output values for 1,000 input value combinations. These MLPs are then fed into the encoders, which output four different MLPs with varying numbers of hidden layers. After feeding the same input values into the decoded MLPs, we obtain the corresponding predictions.

To quantify the effectiveness of the training, we use Median Percentage Error (MPE) as the evaluation metric. The MPE value of the decoded MLP with $i$ hidden layers is defined as:
% $$ MPE_i = Median_x \left( \frac{|\mathcal{N}_s(x) -[H_i(G(\mathcal{N}_s)](x)|}{|\mathcal{N}_s(x)|} \right) $$ 
\begin{equation}
    MPE_i = \text{Median}_x \left( \frac{\left|\mathcal{N}_s(x) - [D_i(E(\mathcal{N}_s))](x)\right|}{\left|\mathcal{N}_s(x)\right|+ \epsilon} \right)
\end{equation}

where $x$ represents the input values, $\mathcal{N}_s(x)$ refers to the output values, and $[D_i(E(\mathcal{N}_s))](x)$ denotes the predictions. We use median value of the percentage error because the magnitude of the ground truth output in the denominator significantly affects the size of the percentage error. If the output is extremely large or small, it can distort the accuracy. A smaller MPE indicates that the decoded MLP more closely resembles the functionality of the input MLP.

% \begin{table}
%   \caption{MPE results for multi-scale autoencoder trained using three different loss functions}
%   \label{MPE_combined_table}
%   \centering
%     \begin{tabular}{lcccc|cccc|cccc}
%         \toprule
%         \multirow{2}{*}{Decoders} & \multicolumn{4}{c|}{MIN Loss} & \multicolumn{4}{c|}{$P=-2$ Loss} & \multicolumn{4}{c}{$P=2$ Loss} \\
%         \cmidrule(lr){2-5} \cmidrule(lr){6-9} \cmidrule(lr){10-13}
%         & $G_1$ & $G_2$ & $G_3$ & $G_4$ & $G_1$ & $G_2$ & $G_3$ & $G_4$ & $G_1$ & $G_2$ & $G_3$ & $G_4$ \\
%         \midrule
%         $H_1$ & \textbf{5.84} & 38.3 & 30.4 & 29.7 & \textbf{8.93} & 34.4 & 36.0 & 43.4 & 31.4 & 38.4 & 37.8 & 45.1 \\
%         $H_2$ & 16.6 & 35.9 & 36.9 & 45.7 & 16.7 & 30.5 & 42.0 & 45.6 & 40.8 & 37.7 & 34.1 & 36.2 \\
%         $H_3$ & 25.4 & \textbf{23.7} & \textbf{24.1} & \textbf{24.4} & 13.1 & \textbf{28.1} & \textbf{25.6} & \textbf{27.3} & \textbf{27.2} & \textbf{35.5} & \textbf{33.1} & \textbf{35.2} \\
%         $H_4$ & 8.37 & 27.5 & 36.0 & 44.7 & 53.4 & 28.5 & 35.1 & 43.2 & 35.0 & 39.9 & 36.2 & 38.2 \\
%         \bottomrule
%     \end{tabular}
% \end{table}

\begin{figure}[h]
    \centering
    \includegraphics[width=\textwidth]{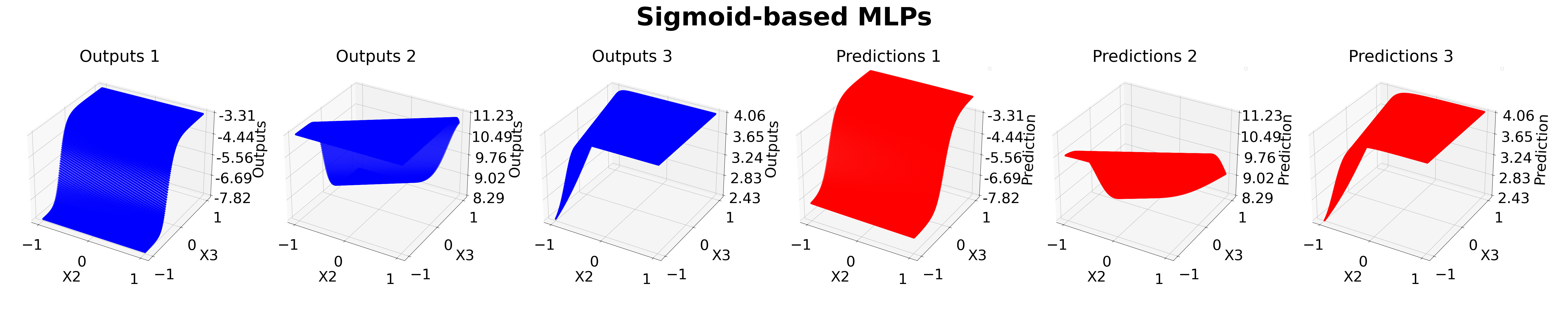}  
    \includegraphics[width=\textwidth]{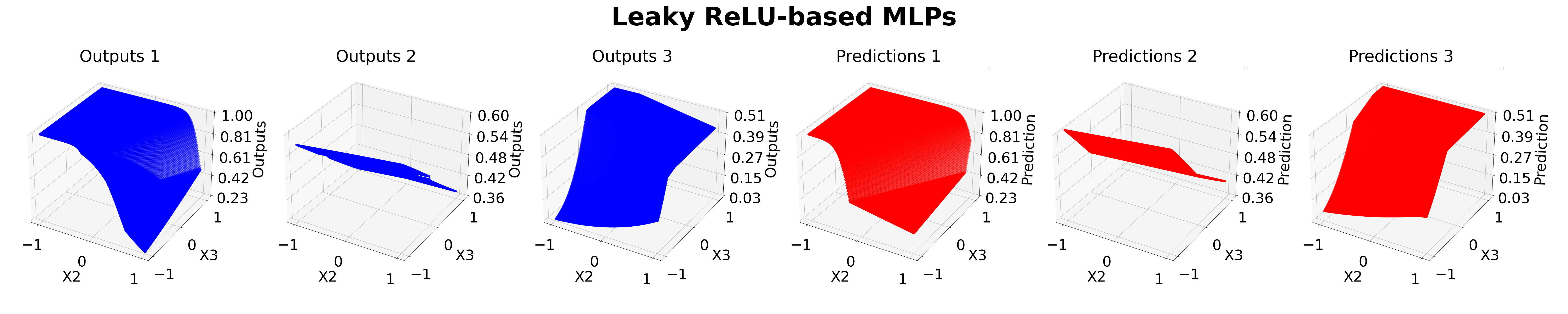} 
    \includegraphics[width=\textwidth]{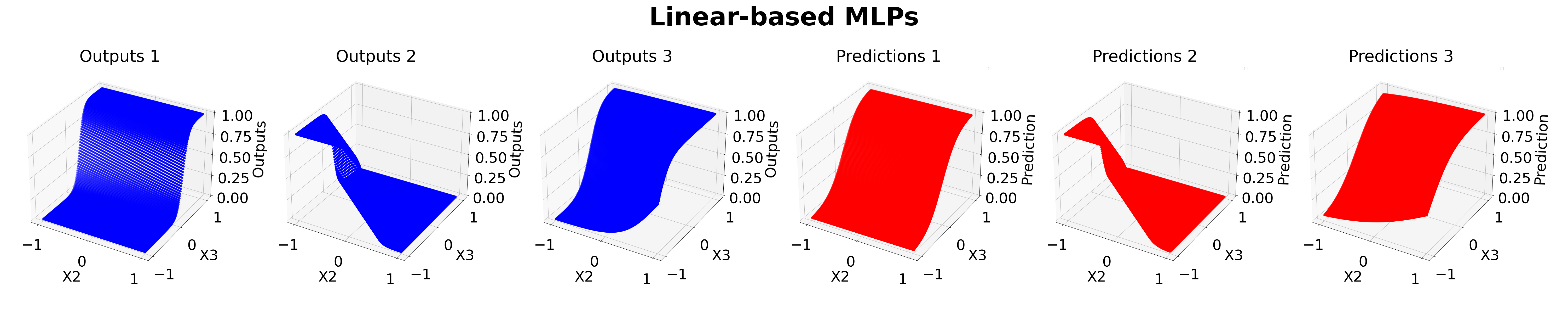}
    \caption{Comparison of functionalities between input MLPs and decoded MLPs using the min loss function. The blue graphs on the left show the outputs of the input MLPs (z-axis as outputs). The red graphs on the right represent predictions of the decoded MLPs with smallest loss over 4 decoders (z-axis as predictions). Since the MLP has three input data points, we fixed one input at 0.5, and the other two inputs were assigned to the x and y axes, ranging in [-1 1].} 
    \label{autoencoder_compare}
\end{figure}

Initially, we train the multi-scale autoencoder using the sigmoid-based MLPs. The performance of the multi-scale autoencoders trained using the three different loss functions is shown in Table~\ref{MPE_combined_table}. The index of each encoder and decoder corresponds to the number of hidden layers in the MLP. From the MPE results, we observe that the autoencoder trained with the $p=2$ loss function has the worst performance. This may be due to the fact that, for some input MLPs, it is difficult to achieve the same level of performance across MLPs with different numbers of hidden layers. In this case, ensuring that at least one of the decoded MLPs can achieve the similar functionality as the input MLP would be a more effective training approach. In the autoencoders trained with $p=-2$ and min loss functions, the 3-hidden-layer decoded MLPs typically perform better. When comparing the best MPE achieved by the four decoders for each input MLP, the model trained with the min loss function performs better. This suggests that additional continuity may not be necessary for effective performance. We also train autoencoders using the min loss function on linear-based and leaky ReLU-based MLP datasets. The specific performance results can be found in the Appendix. To better visualize the effect of autoencoder training, we test 3 sets of MLPs for each activation function, and the comparison plots are shown in Figure\ref{autoencoder_compare}.

\subsection{Searching for Optimal MLPs}
\label{Searching for Optimal MLPs}

% \begin{table}
%   \caption{MPE results of the searched sigmoid MLP across the three datasets}
%   \label{searched_MPE_table}
%   \centering
%     %\begin{tabular}{lccc}
%     \begin{tabular}{lp{1.3cm}p{1.3cm}p{1.3cm}}
%         \toprule
%         Decoders & 2 HL, 0.7 sparsity & 3 HL, 0.5 sparsity & 4 HL, 0.5 sparsity\\
%         \midrule
%         $H_1$     & 9.65  & 19.07  & 24.05        \\
%         $H_2$     & 7.83  & 22.47    & 11.24    \\
%         $H_3$     & 4.34    &  5.83     &  13.32    \\
%         $H_4$     & 15.78    & 5.55   & 13.40     \\
%         \bottomrule
%         \multicolumn{4}{l}{\footnotesize{Note: HL refers to hidden layers.}}
%     \end{tabular}
% \end{table}

\begin{table}[h]
  \caption{MPE and non-zero counts of the searched MLPs across the three datasets for sigmoid-based, leaky ReLU-based, and linear-based MLPs}
  \label{combined_non-zero_table}
  \centering
    \begin{tabular}{lp{1.5cm}p{1.5cm}p{1.5cm}p{1.5cm}}
        \toprule
        MLP Type & Decoders & 2 HL (14) & 3 HL (35) & 4 HL (48) \\
        \midrule
        \multirow{4}{*}{Sigmoid-based} 
        & $D_1$ & 9.65 (\textbf{11})  & 19.07 (\textbf{13})  & 24.05 (\textbf{11}) \\
        & $D_2$ & 7.83 (20)  & 22.47 (\textbf{12})  & \textbf{11.24} (\textbf{10}) \\
        & $D_3$ & \textbf{4.34} (29)  & 5.83 (\textbf{26})  & 13.32 (\textbf{28}) \\
        & $D_4$ & 15.78 (48)  & \textbf{5.55} (40)  & 13.40 (57) \\
        \midrule
        \multirow{4}{*}{Leaky ReLU-based} 
        & $D_1$ & 2.66 (\textbf{4})  & \textbf{10.51} (\textbf{16})  & 12.62 (\textbf{7}) \\
        & $D_2$ & \textbf{0.19} (\textbf{12})  & 22.11 (\textbf{11})  & 0.86 (\textbf{24}) \\
        & $D_3$ & 1.35 (\textbf{11})  & 18.02 (49)  & \textbf{0.35} (\textbf{17}) \\
        & $D_4$ & 0.39 (49)  & 28.27 (\textbf{26})  & 0.50 (\textbf{31}) \\
        \midrule
        \multirow{4}{*}{Linear-based} 
        & $D_1$ & \textbf{3.41} (\textbf{8})  & \textbf{0.01} (\textbf{11})  & 48.05 (\textbf{16}) \\
        & $D_2$ & 3.57 (\textbf{7})  & \textbf{0.01} (\textbf{12})  & \textbf{0.00} (\textbf{17}) \\
        & $D_3$ & 10.73 (43)  & 6.09 (39)  & \textbf{0.00} (\textbf{23}) \\
        & $D_4$ & 5.01 (\textbf{11})  & 0.02 (\textbf{12})  & 0.70 (\textbf{28}) \\
        \bottomrule
    \end{tabular}
    \vspace{0.6em}
    \parbox{\linewidth}{\footnotesize{Note: HL refers to hidden layers. The numbers in parentheses in the table headers represent the non-zero counts of the input MLPs, which are 14, 35, and 48 for the 2, 3, and 4 hidden layers, respectively. Each value in the table follows the format MPE (non-zero count). The MPE values in bold indicate the best performance across four decoders, and the non-zero counts in bold are those that are lower than the non-zero count of the input MLP.}}
\end{table}

\begin{figure}[htp]
    \centering
    \includegraphics[width=\textwidth]{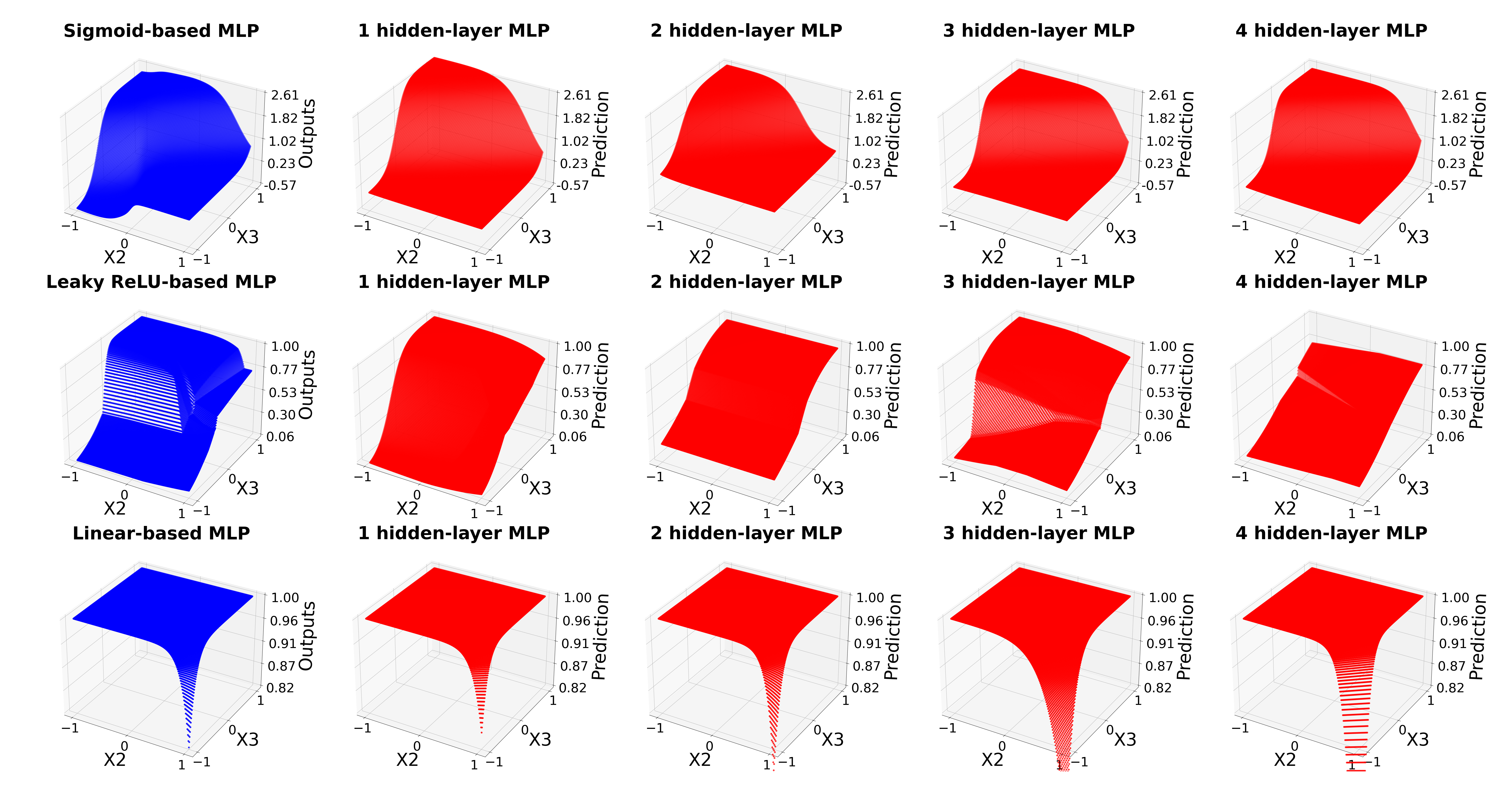}
    \caption{In each row, the dataset is generated by a 3 hidden-layer MLP with 35 non-zero weights for sigmoid-based, leaky ReLU-based, and linear-based MLPs, respectively. The blue plots represent the ground truth outputs, while the red plots from left to right correspond to the 1, 2, 3, and 4 hidden-layer optimal MLPs found through the search.}
    \label{3_Hidden_combinations}
\end{figure}

\begin{figure}[h!]
    \centering
    \begin{minipage}[b]{0.32\textwidth}
        \centering
        \includegraphics[width=\textwidth]{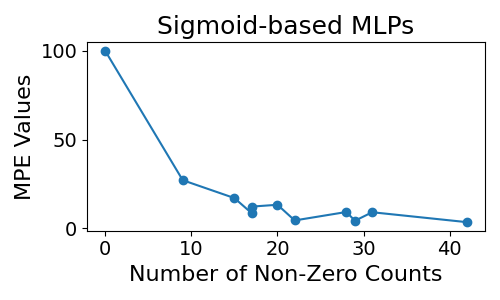}
    \end{minipage}
    \hspace{0.001\textwidth} 
    \begin{minipage}[b]{0.32\textwidth}
        \centering
        \includegraphics[width=\textwidth]{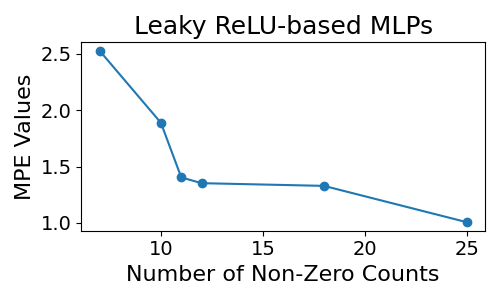}
    \end{minipage}
    \hspace{0.001\textwidth}
    \begin{minipage}[b]{0.32\textwidth}
        \centering
        \includegraphics[width=\textwidth]{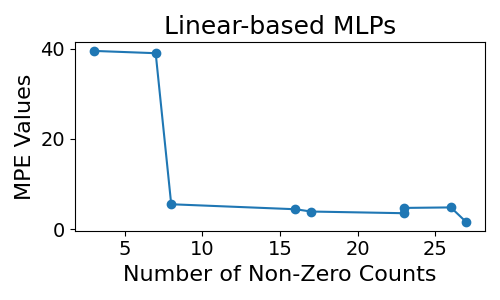}
    \end{minipage}
    \caption{This figure displays three plots generated using datasets from a 2 hidden-layer MLP with 14 non-zero weights and Decoder 3. The non-zero count was adjusted by modifying the weight of the sparsity term in the loss function.}
    \label{fig:three_count_MPE}
\end{figure}

%%%%%Mansooreh

Since we found that the autoencoder trained with the min loss function performed the best, we use this autoencoder in the search for the optimal MLP. We conduct the search using datasets generated by three MLPs for each type. Each set of MLPs consists of networks with 2, 3, and 4 hidden layers, and non-zero counts of 14, 35, 48, respectively. These non-zero counts represent the number of active weights in each MLP, which we use as a measure of compactness. The fewer the non-zero elements, the more compact the MLP becomes. All hidden layers in the MLPs used to generate the datasets contained 5 neurons. Each dataset consists of 100,000 input-output pairs, which we split into training, validation, and test sets in a 5:3:2 ratio. We fix the parameters in the autoencoder and perform gradient descent in the embedding space using the loss function with sparsity penalty, aiming to find a sparse MLP that performs well on the dataset. For each dataset, we conduct a parallel search for the optimal MLPs with 1, 2, 3, and 4 hidden layers. 

The specific results are shown in Tables~\ref{combined_non-zero_table}. From the sigmoid-based MPE data, It is clear that 3 hidden-layer MLPs are often the most effective. This could be because a 3 hidden-layer sigmoid-based MLP acts as a universal model, capable of simulating both deeper networks and simplifying shallower networks. Additionally, as noted in Section \ref{multi-scale training} (see "Multi-Scale Encoder-Decoder"), the performance of Decoder 3 was particularly strong during training the multi-scale autoencoder for sigmoid-based MLPs, which may explain why using Decoder 3 results in better sigmoid-based MLPs. Conversely, using a 1 hidden-layer sigmoid-based MLP to simulate deeper networks proves to be challenging, since the performance of Decoder 1 declines as the depth of the MLPs increases. From the non-zero counts data, the sparsity penalty in our loss function effectively reduces the number of neurons involved in computations within complex networks. However, further simplifying networks that are already very sparse and compact proves to be quite challenging, as for the 2 hidden-layer sigmoid-based MLP with 14 non-zero weights, only the searched MLP with 1 hidden-layer can be more compact than it. For the linear-based and leaky ReLU-based MLP datasets, using a 1 hidden-layer MLP to represent a 4 hidden-layer dataset proved challenging, as Decoder 1 did not perform well in searching for MLPs in these datasets. Additionally, shallower MLPs are better at improving compactness and sparsity, as the smallest non-zero counts typically appear in the 1-layer or 2-layer searched MLPs. To better visualize and understand the performance of the searched MLPs, we generate 3D plots shown in Figure ~\ref{3_Hidden_combinations}. The output shapes of the input MLPs and the prediction shapes of the decoded MLPs are similar, indicating comparable responses to the same inputs.

From the plot of the number of non-zero counts versus MPE values in Figure ~\ref{fig:three_count_MPE}, it can be seen that as the number of non-zero counts increases, the MPE tends to decrease. This indicates that there is a tradeoff between sparsity, compactness, and model performance. However, in the early stages of reducing the non-zero count, the increase in MPE is relatively slow. Therefore, we can identify a region where the increase in MPE is slower and find the minimum non-zero count that can be achieved within this region to balance sparsity, compactness, and performance.

\section{Conclusion and Future Work}

We proposed a multi-scale encoder-decoder structure to encode neural networks and form a continuous embedding space. By performing gradient descent in this low-dimensional space, we simultaneously optimized the network's architecture and weights for any given dataset. We demonstrated that this framework is effective for searching sigmoid-based, leaky ReLU-based, and linear-based MLPs. For future work, we will expand the types and complexities of candidate neural networks, potentially including Temporal Convolutional Networks (TCNs), Convolutional Neural Networks (CNNs), and Recurrent Neural Networks (RNNs), and embed these different architectures into a unified embedding space to facilitate the search for diverse types of networks.

% [1] Alexander, J.A.\ \& Mozer, M.C.\ (1995) Template-based algorithms for
% connectionist rule extraction. In G.\ Tesauro, D.S.\ Touretzky and T.K.\ Leen
% (eds.), {\it Advances in Neural Information Processing Systems 7},
% pp.\ 609--616. Cambridge, MA: MIT Press.

\bibliographystyle{plainnat}
\bibliography{preprint} % This should match the name of your .bib file

\clearpage
\section*{Supplementary Material}

\paragraph{Random generation process for the input MLPs} 

Here is the process of generating our input MLPs, which are used to train the multi-scale autoencoder and to create datasets for finding the optimal MLP:

\begin{enumerate}
    \item Randomly choose $l \in \{1, 2, 3, 4\}$ for the number of hidden layers.
    
    \item For each network, generate sparsity patterns by randomly removing 50\%, 70\%, 80\%, and 90\% of the links, ensuring at least one path exists from each input to the output.
    \begin{itemize}
        \item Ensure connectivity by randomly selecting and fixing one path from each input to the output, then prune the remaining links.
    \end{itemize}
    
    \item Assign random weights between [-10, 10] for sigmoid-based MLP, [-3, 3] for both leaky ReLU-based and linear-based MLPs to the remaining edges for each sparsity pattern.
\end{enumerate}

\paragraph{Algorithm for training multi-scale autoencoder and searching optimal MLPs}

Algorithm \ref{algo1} gives overview of the two key phases: (1) training the multi-scale autoencoder and (2) searching for the optimal MLPs. 

\begin{algorithm}
\caption{Simultaneous Weight and Architecture Optimization for Neural Networks}
\label{algo1}
\begin{algorithmic}[1]
\Require For training the multi-scale autoencoder, we need input and output value pairs $(X, Y)$, the input MLP $M$, and hyperparameters such as batch size $b$, number of epochs $e$, and learning rate $\lambda_1$. For training the best MLP over the dataset, we need the dataset's input and output pairs $(X_d, Y_d)$ for calculating loss, number of iterations $R$, the learning rate $\lambda_2$ for embedding $z$, and $\lambda_3$ for threshold $t$.
\Ensure the searched neural network $M'$ is sparse and compact while achieving optimal performance on the given dataset
\Statex
\Procedure{Training multi-scale autoencoder}{}
    \For{$Epoch = 1, \dots, e$} 
        \State Split the input MLPs $M$ and its corresponding $(X, Y)$ pairs into batches of size $b$.
        \For{each batch}
            \State $\phi \gets \phi - \lambda_1 \frac{\partial L_{ED}}{\partial \phi}$ \Comment{Update encoders $E$ parameters}
            \State $\theta \gets \theta - \lambda_1 \frac{\partial L_{ED}}{\partial \theta}$ \Comment{Update decoders $D$ parameters}
        \EndFor
    \EndFor
    \State \textbf{return} the optimized $E$, $D$
\EndProcedure
\Statex
\Procedure{Training optimal MLP}{}
    \For{$i = 1, \dots, I$} \Comment{Parallelly find $I$ optimal MLPs with different hidden layers}
        \State Sample a point from the embedding space $z_i \in Z$, initialize $t_i$ = 0.
        \For{$r = 1, \dots, R$}
            \State $z_i \gets z_i - \lambda_2 \frac{\partial L_i}{\partial z_i}$ \Comment{Update embedding $z_i$ by gradient descent}
            \State $t_i \gets t_i - \lambda_3 \frac{\partial S_i(z_i)}{\partial t_i}$ \Comment{Update threshold $t_i$ by gradient descent}
        \EndFor
        \State Decode each $z_i$ to architecture $M_i' = D_i(z_i)$ \Comment{Decode embedding to an MLP}
    \EndFor
    \State \textbf{return} the discovered architectures $\{M_1', M_2', \dots, M_I'\}$
\EndProcedure

\end{algorithmic}
\end{algorithm}

\paragraph{Training Multi-scale Autoencoder on Leaky ReLU-based and Linear-based MLPs}

As referenced in Section \ref{multi-scale training}, using min loss function \ref{min loss} during the training process yielded the best performance. Consequently, we also trained autoencoders using the min loss function for both leaky ReLU-based and Linear-based MLPs. The specific MPE results for these experiments can be found in the table \ref{LR and LN MPE_combined_MLP_table}. For each MLP type, the MPE is measured across four decoders, and the best-performing results are bolded.

\begin{table}[htbp]
  \caption{MPEs for multi-scale autoencoder trained on leaky ReLU-based and linear-based MLPs}
  \label{LR and LN MPE_combined_MLP_table}
  \centering
    \begin{tabular}{lcccc|cccc}
        \toprule
        \multirow{2}{*}{Decoders} & \multicolumn{4}{c|}{LeakyReLU-based MLP} & \multicolumn{4}{c}{Linear-based MLP} \\
        \cmidrule(lr){2-5} \cmidrule(lr){6-9}
        & $E_1$ & $E_2$ & $E_3$ & $E_4$ & $E_1$ & $E_2$ & $E_3$ & $E_4$ \\
        \midrule
        $D_1$ & \textbf{6.05}  & \textbf{11.90} & \textbf{9.44}  & \textbf{12.60} & 41.99 & \textbf{47.36} & 63.95 & 81.22 \\
        $D_2$ & 7.38  & 11.94 & 9.87  & 14.93 & 42.72 & 59.00 & 60.42 & \textbf{65.81} \\
        $D_3$ & 9.20  & 12.40 & 9.69  & 18.20 & 42.76 & 54.74 & 68.75 & 69.88 \\
        $D_4$ & 8.60  & 15.36 & 14.33 & 28.20 & \textbf{39.00} & 59.83 & \textbf{58.91} & 73.30 \\
        \bottomrule
    \end{tabular}
\end{table}

\paragraph{Visualize optimal searched MLP over datasets}

\begin{figure}[htp]
    \centering
    \includegraphics[width=\textwidth]{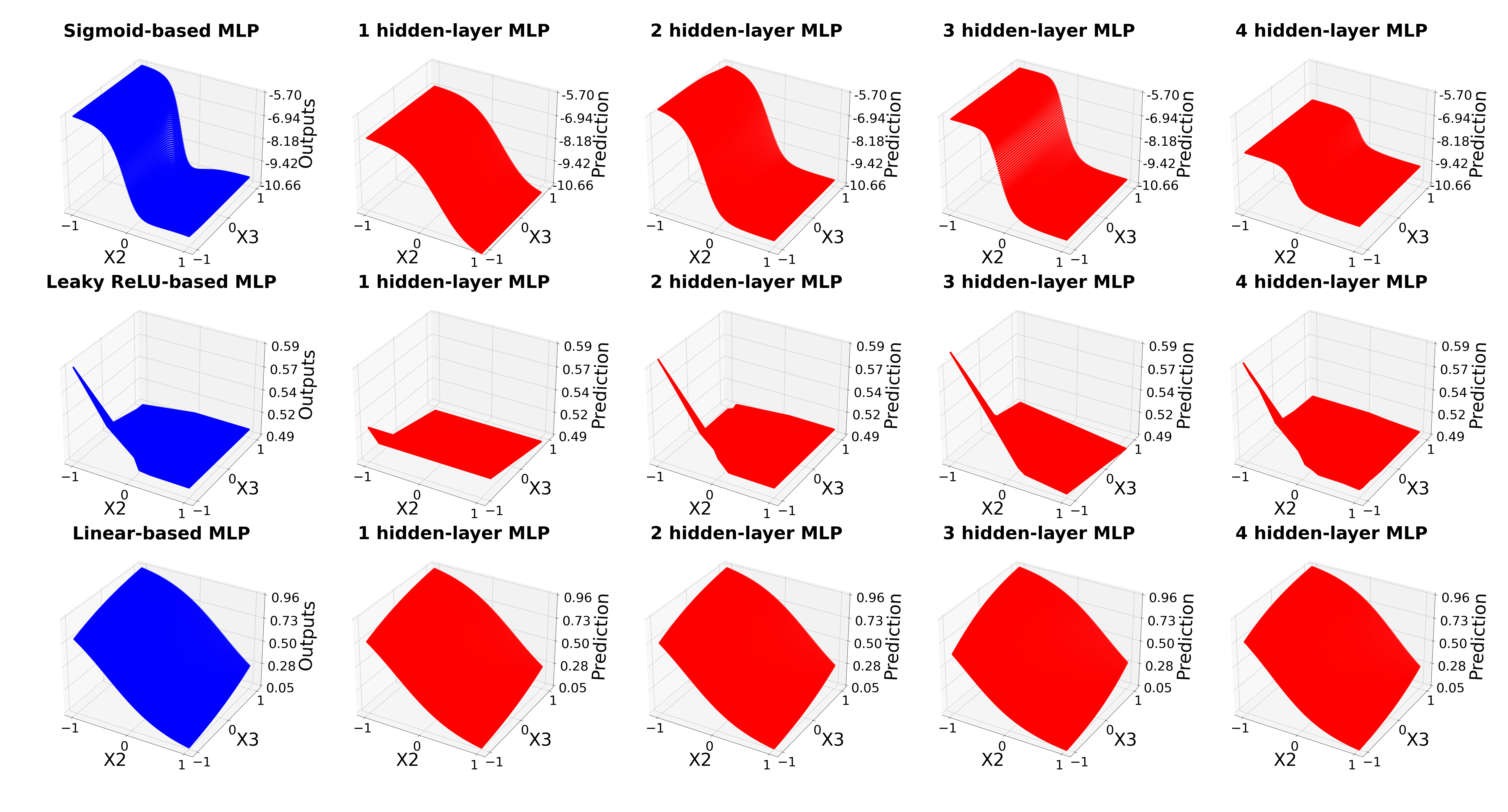}
    \caption{In each row, the dataset is generated by a 2 hidden-layer MLP with 14 non-zero weights for sigmoid-based, leaky ReLU-based, and linear-based MLPs, respectively.}
    \label{2_Hidden_combinations}
\end{figure}

\begin{figure}[htp]
    \centering
    \includegraphics[width=\textwidth]{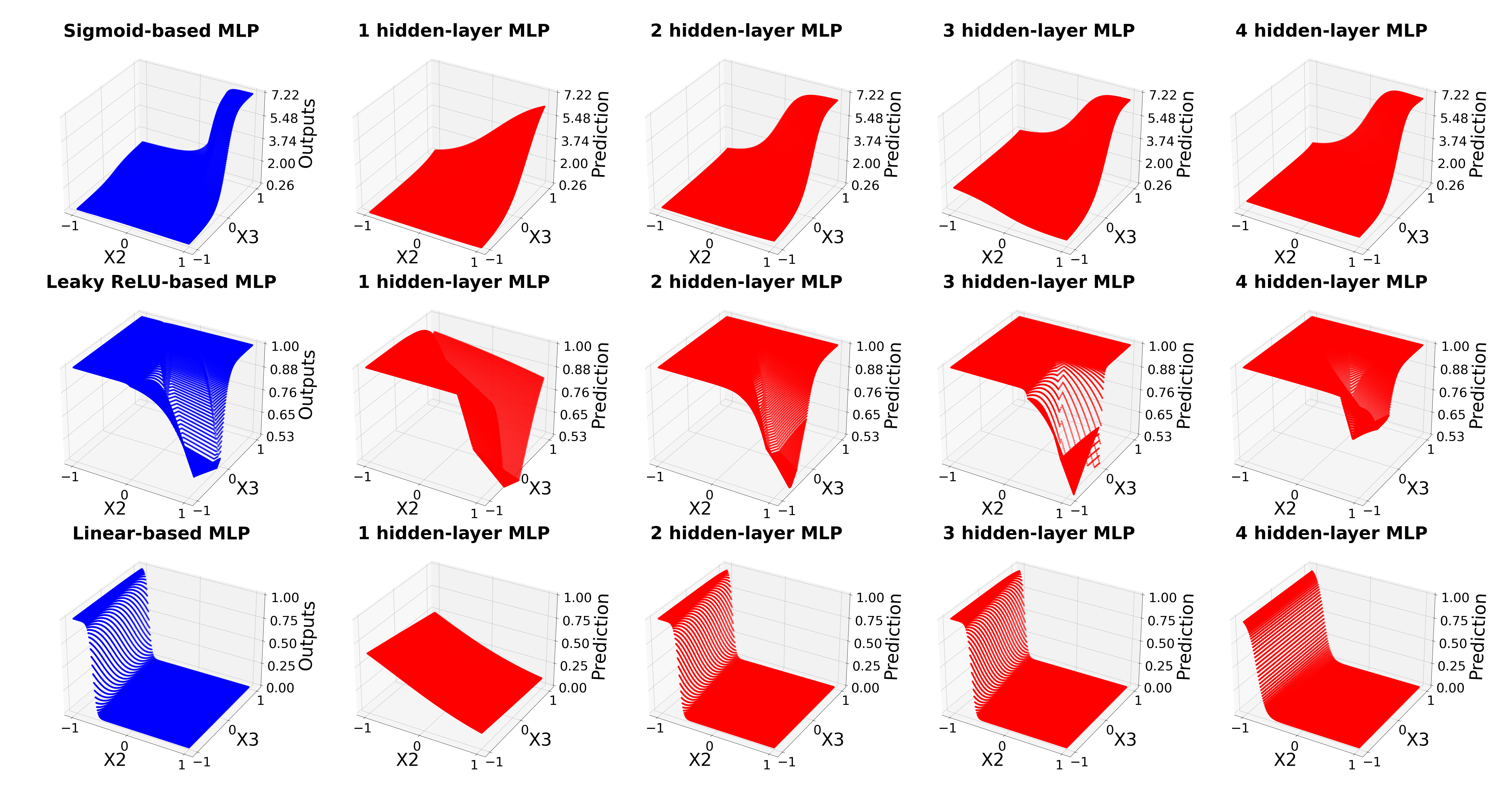}
    \caption{In each row, the dataset is generated by a 4 hidden-layer MLP with 48 noon-zero weights for sigmoid-based, leaky ReLU-based, and linear-based MLPs, respectively.}
    \label{4_Hidden_combinations}
\end{figure}

Similar as what we did in Section \ref{Searching for Optimal MLPs}, we applied the method to search for the optimal MLP using datasets generated by 2 hidden-layer and 4 hidden-layer MLPs with 14 and 48 non-zero weights, respectively. The visualization of the optimal MLP search process for these datasets can be seen in Figures \ref{2_Hidden_combinations} and \ref{4_Hidden_combinations}.

% {\color{red}

% Autoencoder experiments (all with masking)

% - input vs output MPE: Tables showing  [sigmoid, ReLU, Linear] x [p = 2, -2, min] x [1-4, 1-4]

% - Input and output 3D plots: They mostly match

% Training MLPs with Autoencoder

% - MPE: Tables showing  [sigmoid, ReLU, Linear] x [min] x [1-4, 1-4], also show sparsity

% - Input and output 3D plots: They mostly match

% - MPE vs sparsity plots

% }

\end{document}